\documentclass{new_tlp}

\usepackage{graphicx}

\usepackage{amsmath}
\usepackage{amssymb,amsfonts}
\usepackage{url}

\usepackage{algorithm}	
\usepackage{algorithmic}

\title[Reasoning about Complex Networks: A Logic Programming Approach]{Reasoning about Complex Networks: \\
A Logic Programming Approach}

\author[P.\ Shakarian, G.I.\ Simari, and D.\ Callahan]{PAULO SHAKARIAN $^1$ \ \ \ \ \ GERARDO I.\ SIMARI $^2$ \ \ \ \ \ DEVON CALLAHAN $^1$ \\
$^1$ Network Science Center and Dept.\ of Electrical Engineering and Computer Science \\
U.S.\ Military Academy, West Point, NY 10996, USA \\
$^2$ Department of Computer Science, University of Oxford \\
Wolfson Building, Parks Road, Oxford OX1 3QD, United Kingdom \\
\email{paulo@shakarian.net, gerardo.simari@cs.ox.ac.uk, devon.callahan@usma.edu}}

\jdate{June 2013}
\pubyear{2013}
\pagerange{\pageref{firstpage}--\pageref{lastpage}}

\newtheorem{example}{Example}[section]
\newtheorem{definition}{Definition}[section]
\newtheorem{theorem}{Theorem}[section]
\newtheorem{lemma}{Lemma}[section]

\def\smallblacksq{{\tiny $\blacksquare$}}

\def\true{\textsf{True}}

\def\dinm{{d_{*}^{in}}}
\def\FB{\mathit{FBnd}}
\def\RB{\mathit{RBnd}}
\def\IB{\mathit{IBnd}}
\def\ELIG{\mathit{Elig}}
\def\QUAL{\mathit{Qual}}

\def\BOUND{\mathit{Bound}}
\def\true{\textsf{Tr}}
\def\false{\textsf{F}}
\def\bnd{\mathit{bnd}}
\def\strtie{\textit{strTie}}
\def\wktie{\textit{wkTie}}
\def\male{\textit{male}}
\def\female{\textit{fem}}
\def\watchesA{\textit{visPgA}}
\def\watchesB{\textit{visPgB}}

\def\gedge{g_{edge}}
\def\gnode{g_{node}}
\def\mancalog{\textsf{MANCaLog}}

\def\softtip{{\mathit{sftTp}}}

\def\negtip{{\mathit{ngTp}}}

\def\ifl{\mathit{ifl}}

\def\IFL{\mathit{IFL}}

\def\setN{\textbf{N}}
\def\calf{\mathcal{F}}

\def\tarrow{\stackrel{\Delta t}{\leftarrow}}

\def\thrarrow{\stackrel{3}{\leftarrow}}

\def\twoarrow{\stackrel{2}{\leftarrow}}
\def\onearrow{\stackrel{1}{\leftarrow}}

\def\call{\mathcal{L}}
\def\calg{\mathcal{G}}

\def\cali{\mathcal{I}}

\def\tm{t_\mathit{max}}

\def\<{\langle}
\def\>{\rangle}

\def\call{\mathcal{L}}

\def\calf{\mathcal{F}}

\begin{document}

\label{firstpage}

\maketitle

\begin{abstract}
Reasoning about complex networks has in recent years become an important topic of study due to its many applications: the adoption of commercial products, spread of disease, the diffusion of an idea, etc. In this paper, we present the MANCaLog language, a formalism based on logic programming that satisfies a set of desiderata proposed in previous work as recommendations for the development of approaches to reasoning in complex networks. To the best of our knowledge, this is the first formalism that satisfies all such criteria. We first focus on algorithms for finding minimal models (on which multi-attribute analysis can be done), and then on how this formalism can be applied in certain real world scenarios. Towards this end, we study the problem of deciding
group membership in social networks: given a social network and a set of groups where group membership of only some of the individuals in the network is known, we wish to determine a degree of membership for the remaining group-individual pairs. 
We develop a prototype implementation
that we use to obtain experimental results on two real world datasets, including a current social network of criminal gangs in a major U.S.\ city.  We then show how the assignment of degree of membership to nodes in this case allows for a better understanding of the criminal gang problem when combined with other social network mining techniques---including detection of sub-groups and identification of core group members---which would not be possible without further identification of additional group members.
\end{abstract}



\section{Introduction and Related Work}
\label{sec:intro}

An epidemic working through a population, cascading electrical power failures, product adoption, and the spread of a mutant gene are all examples of diffusion processes that can happen in complex networks.  These network processes have been studied in a variety of disciplines, including computer science~\cite{kleinberg}, biology~\cite{liebermanEvolutionary2005}, sociology~\cite{Gran78}, economics~\cite{schelling}, and physics~\cite{sood08}.  Much existing work in this area is based on pre-existing models in sociology and economics---in particular the work of~\cite{Gran78,schelling}.  However, recent examinations of social networks---both analysis of large data sets and observational studies---have indicated that there may be additional factors to consider that are not taken into account by these models.  These include the attributes of nodes and edges, competing diffusion processes, and time.  In this paper, we propose \mancalog\ ({\em Multi-Attribute Networks and Cascades}), a logical language for
modeling multi-attribute processes in complex networks that can richly express how individuals in the
network adopt or fail to adopt certain behaviors, and how these behaviors diffuse through the network.
\mancalog\ is based on a set of design criteria recently proposed in~\cite{mancalogAAMAS13}, and it is
to the best of our knowledge the first logical language for modeling diffusion in complex networks that meets these criteria.
We also introduce fixed-point based algorithms for computing the result of a diffusion process.  Note that these algorithms are proven not only to be correct, but also to run in polynomial time.  Hence, our approach can not only better express many aspects of multi-attribute processes in complex networks, but it can do so in a reasonable amount of time.
Finally, we investigate applications by considering the problem of deciding
group membership in social networks: given a social network and a set of groups where  membership of only some of the individuals is known, we wish to determine a degree of membership for the remaining group-individual pairs. We also develop a prototype implementation
that we use to obtain experimental results on two real world datasets, including a current social network of criminal gangs in a major U.S.\ city.

\subsection{Design Criteria}

In recent work~\cite{mancalogAAMAS13}, we proposed a set of seven design criteria that we believe
a framework for reasoning about multi-attribute processes in complex networks should satisfy.
As a quick overview, these criteria are:
(i) Multiply labeled and weighted nodes and edges:
Many existing frameworks for studying diffusion in complex networks assume that there is only one type of vertex that may become ``active'' or may ``mutate'' and only one possible relationship between nodes;
however, in reality nodes and edges often have different properties.  For instance, labels on edges can be used to differentiate between strong and weak ties (edge types);
(ii) Explicit representation of time:
Most work in the literature either assumes static models or makes several simplifying assumptions such as a model of time solely based on temporal decay of influence; we seek a richer model of temporal relationships between conditions in the network structure, the current state of the cascades in process,
and how influence propagates;
(iii) Non-Markovian temporal relationships:
Temporal dependencies should be able to span multiple units of time; hence, the ``memoryless'' mode of a standard Markov process is insufficient.  We strive to create a framework where dependencies can be from other earlier time steps;
(iv) Representation of uncertainty:
In practice, it is not always possible to judge the attributes of all individuals in a network, and thus an element of uncertainty must be included. In connection with point~(vii) below, this should not be at the expense of tractability;
(v) Competing processes:
Real-world situations often present competing network processes, where the success of one hinges on the failure of the other;
(vi) Non-Monotonic Processes:
Though in much existing work on diffusion processes in complex networks the number of nodes attaining a certain property at each time step can only increase, if we allow for competing cascades in the same model we cannot have such a strong restriction; and
(vii) Tractability:
The social networks of interest in today's data mining problems often have millions of nodes, and it is reasonable to expect that soon billion-node networks will be commonplace.  Any framework for dealing with these problems must be tractable and offer areas for practical improvement for further scalability.

\subsection{Related Work}
\label{sec:relwork}

The above criteria can be summarized as the desire to design the most expressive language for network cascades possible while still allowing computation of the outcome of a diffusion process to be completed in a tractable amount of time.  As a comparison, let us briefly describe some relevant related work.  Perhaps the best known general model for representing diffusion in complex networks is the independent cascade/linear threshold (IC/LT) model of~\cite{kleinberg}. However, although this framework was shown to be capable of expressing a wide variety of sociological models, it assumes the Markov property and does not allow for the representation of multiple attributes on vertices and edges. A more recent framework, social network optimization problems (SNOPs)~\cite{snops} uses logic programming to allow for the representation of attributes, but this framework does not allow for competing processes or non-monotonic cascades. A related logic programming framework, competitive diffusion (CD)~\cite{bss10} allows for competitive diffusion and non-monotonic processes but does not explicitly represent time and also makes Markovian assumptions. Further, we also note that the semantics of CD yields a ``most probable interpretation'' that is not a unique solution. Hence, a given model in that framework can lead to multiple and possibly contradictory, outcomes to a cascade (this problem is avoided in $\mancalog$).  Another popular class of models is Evolutionary Graph Theory (EGT)~\cite{liebermanEvolutionary2005}, which is highly related to the voter model (VM)~\cite{sood08}. Although this framework allows for competing processes and non-monotonic diffusion, it also makes Markovian assumptions while not explicitly representing time.  Further, determining the outcome of a cascade in those models is NP-hard, while determining the outcome in $\mancalog$ can be accomplished in polynomial time.  Table~\ref{rwTab} lists how these models compare to $\mancalog$ when considering our design criteria.

\begin{figure}
\footnotesize
\centering
\begin{tabular}{lccccc}\hline
Criterion  & MANCaLog &  IC/LT & SNOP & CD & EGT/VM \\ \hline
1.\ Labels & Yes &No & Yes &Yes &No  \\
2.\ Explicit Representation of Time & Yes &  No & Yes & No & Yes  \\
3.\ Non-Markovian Time & Yes &No & No &No & No \\
4.\ Uncertainty & Yes &   Yes & Yes &Yes & Yes  \\
5.\ Competing Processes & Yes & No & No &Yes &Yes  \\
6.\ Non-monotonic Processes & Yes &No & No &Yes& Yes \\
7.\ Tractablity & PTIME & $\#$P-hard & PTIME & PTIME & NP-hard  \\ \hline
\end{tabular}
\caption{A comparison of models.}
\label{rwTab}
\end{figure}

\medskip

The rest of this paper is organized as follows: Section~\ref{sec:mancalog} presents the
\mancalog\ framework; Section~\ref{sec:cons-ent-fix} discusses consistency, entailment, and
fixpoint computation of minimal models; Section~\ref{sec:applications} discusses applications in
social networks and experimental results, and Section~\ref{sec:conc} includes conclusions and
future work.


\section{The \mancalog\ Language: Syntax and Semantics}
\label{sec:mancalog}


In this work we assume that individuals (persons, agents, etc.) are arranged in a directed graph (or network) $G=(V,E)$, where the set of nodes corresponds to the individuals, and the edges model the relationships between them.
We also assume a set of labels $\call$, which is partitioned into two sets: {\em fluent} labels $\call_{f}$ (labels that can change over time) and {\em non-fluent} labels $\call_{nf}$ (labels that do not); labels can be applied to both the nodes and edges of the network.  We will use the notation $\calg = V \cup E$ to be the set of all \textit{components} (nodes and edges) in the network.  Thus, $c \in \calg$ could be either a node or an edge.

\begin{figure}[t!]
\centering
\includegraphics[width=0.45\textwidth]{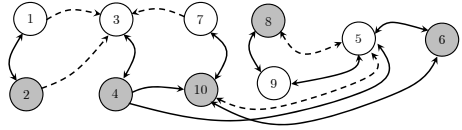}
\caption{Simple online social network $G_{soc}$.  Solid edges are labeled with $\strtie$ and dashed edges with $\wktie$.  White nodes are labeled with $\male$ and gray nodes with $\female$. Arrows represent the direction of the edge; double-headed edges represent two edges with the same label.}
\label{nwFig}
\end{figure}

\begin{example}
\label{ex1}
We will use the sample online social network $G_{soc}$ shown in Figure~\ref{nwFig} as the running example; $\calg_{soc}$ is used to denote the set of components of $G_{soc}$.  Here we have $\call_{nf}=\{\male,\female,$ $\strtie,\wktie\}$ representing male, female, strong ties and weak ties, respectively.  Additionally, we have $\call_{f} = \{\watchesA,\watchesB \}$ representing visiting webpage A and visiting webpage B, respectively.
\hfill\smallblacksq
\end{example}

We now present a logical language where we use atoms, referring to labels and weights, to describe properties of the nodes and edges.  Though labels themselves could be modeled as atoms instead of predicates (to model non-ground labelings that allow for greater expressibility), for simplicity of presentation we leave this to future work.  The first piece of the syntax is the \textbf{network atom}.

\begin{definition}[Network Atom]
Given label $L \in \call$ and real-valued interval $\bnd \subseteq [0,1]$ (referred to as a ``weight interval''), a \textit{network atom} is of the form $\langle L, \bnd \rangle$.  A network atom is \textit{fluent} (resp., \textit{non-fluent}) if $L \in \call_{f}$ (resp., $L \in \call_{\mathit{nf}}$). The set of possible network atoms is denoted with $\mathit{NA}$.
\end{definition}

Network atoms describe  properties of nodes and edges.  The definition is intuitive: $L$ represents a property of the vertex or edge, and associated with this property is some weight that may have associated uncertainty---hence represented as an interval $\bnd$, which can be open or closed. An invalid bound is represented by $\emptyset$, which is equivalent to all other invalid bounds.

\begin{definition}[World]
A \textit{world} $W$ is a set of network atoms such that for each $L \in \call$ there is no more than one network atom of the form $\<L,\bnd \>$ (where $\bnd \neq \emptyset$) in $W$.
\end{definition}

A \textbf{network formula} over $NA$ is defined using conjunction, disjunction, and negation in the usual way.  If a formula contains only non-fluent (resp., fluent) atoms, it is a non-fluent (resp., fluent) formula.

\begin{definition}[Satisfaction of Worlds]
Given world $W$ and network formula $f$, \textit{satisfaction} of $W$ by $f$ is defined as follows:

\noindent
$-$ If $f = \langle L, [0,1]  \rangle$ then $W \models f$.

\noindent
$-$ If $f = \langle L, \emptyset  \rangle$ then $W \not\models f$.

\noindent
$-$ If $f = \langle L, \bnd \rangle$, with $\bnd \neq \emptyset$ and $\bnd \neq [0,1]$, then
$W \models f$ iff there exists $\langle L, \bnd' \rangle \in W$ s.t. $\bnd' \subseteq \bnd$.

\noindent
$-$  If $f = \neg f'$ then $W \models f$ iff $W \not\models f'$.

\noindent
$-$ If $f = f_1 \wedge f_2$ then $W \models f$ iff $W \models f_1$ and $W \models f_2$.

\noindent
$-$ If $f = f_1 \vee f_2$ then $W \models f$ iff $W \models f_1$ or $W \models f_2$.
\end{definition}

For some arbitrary label $L \in \call$, we will use the notation $\true = \<L,[0,1]\>$ and $\false = \<L,\emptyset \>$ to represent a tautology and contradiction, respectively.  For ease of notation (and without loss of generality), we say that if there does not exist some $\bnd$ s.t.\ $\<L, \bnd\> \in W$, then this implies that $\<L, [0,1]\> \in W$.

\begin{example}
\label{ex2}
Following from Example~\ref{ex1}, the network atom $\< female, [1,1] \>$ can be used to identify a node as a woman.
World $W =
\{\<\female, [1,1]\>, \<\male, [0,0]\>, \<\watchesA,[1,1]\>, \<\watchesB,[0,0]\>\}$
might be used to identify a woman who visits webpage A.  Clearly, we have that
$W \models
\< \female, [1,1] \> \wedge \neg \<\watchesA,[0.5,0.9]\> \wedge \neg \<\watchesB,[0.1,0.7]\>$.
\hfill\smallblacksq
\end{example}

The idea is to use $\mancalog$ to describe how properties (specified by labels) of the nodes in the network change over time.  We assume that there is some natural number $\tm$ that specifies the total amount of time we are considering, and we use $\tau = \{t \; | \; t \in [0,\tm]\}$ to denote the set of all time points.  How well a certain property can be attributed to a node is based on a \textit{weight} (to which the $\bnd$ bound in the network atom refers).  As time progresses, a weight can either increase/decrease and/or become more/less certain. We now introduce the $\mancalog$ fact, which states that some network atom is true for a node or edge during certain times.

\begin{definition}[$\mancalog$ Fact]
If $[t_1,t_2] \subseteq [0,\tm]$, $c \in \calg$, and $a \in NA$, then $(a,c):[t_1,t_2]$ is a $\mancalog$ \textbf{fact}.  A fact is \textit{fluent} (resp., \textit{non-fluent}) if atom $a$ is fluent (resp., non-fluent).  All non-fluent facts must be of the form $(a,c):[0,\tm]$.  Let $\calf$ be the set of all facts and $\calf_{nf}, \calf_{f}$ be the set of all non-fluent and fluent facts,
respectively.
\end{definition}
An example of a fact based on the running example is $F = (\<\male,[1,1]\>,1):[0,\tm]$.
Next, we introduce integrity constraints (ICs).

\begin{definition}[Integrity constraint]
Given fluent network atom $a$ and conjunction of network atoms $b$, an \textit{integrity constraint}
is of the form $a \hookleftarrow b$.
\end{definition}

Intuitively, integrity constraint $\<L,\bnd\> \hookleftarrow b$ means that if at a certain time point a component (vertex or edge) of the network has a set of properties specified by conjunction $b$, then at that same time the component's weight for label $L$ must be in interval $\bnd$.
Following from the previous examples, the integrity constraint $\<\male,[0,0]\> \hookleftarrow \<\female,[1,1]\>$ would require any node designated as a female to not be male.

We now turn to $\mancalog$ rules.  The idea behind rules is simple: a node that meets some criteria is influenced by the set
of its neighbors who possess certain properties.  The amount of influence exerted on a node by its neighbors is specified by an \textit{influence function}, whose precise effects will be described later on when we discuss the semantics.
As a result, a rule consists of four major parts: (i) an influence function, (ii) neighbor criteria, (iii) target criteria, and (iv) a target.  Intuitively, (i) specifies how the neighbors influence the node in question, (ii) specifies which of the neighbors can influence the node, (iii) specifies the criteria that cause the node to be influenced, and (iv) is the property of the node that changes as a result of the influence.

We will discuss each of these parts in turn, and then define rules in terms of these elements.  First, we define influence functions and neighbor criteria.

\begin{definition}[Influence Function]
An \textit{influence function} is a function $\ifl : \setN \times \setN \rightarrow [0,1] \times [0,1]$ that satisfies
the following two axioms:

\smallskip
\noindent
1.\ $\ifl$ can be computed in constant ($O(1)$) time.

\smallskip
\noindent
2.\ For $x'>x$ we have $\ifl(x',y) \subseteq \ifl(x,y)$.

\smallskip
\noindent
We use $\IFL$ to denote the set of all influence functions.
\end{definition}

Intuitively, an influence function takes the number of qualifying influencers (those that meet some requirement to be able to influence a certain individual, yet may or may not carry a contagion) and the number of eligible influencers (those that meet some requirement to be able to influence a certain individual \textit{and} carry a contagion) and returns a bound on the new value for the weight of the property of the target node that changes.
In practice, we expect the time complexity of such a function to be polynomial in terms of its arguments. However, as both arguments are naturals bounded by the maximum degree of a node in the network, this value will be much smaller than the size of the network---we thus treat it as a constant here.

\begin{definition}[Neighbor Criterion]
If $\gedge,\gnode$ are non-fluent network formulas, $h$ is a conjunction of network atoms, and $\ifl$ is an influence function, then
$
(\gedge,\gnode,h)_\ifl
$
is a \textit{neighbor criterion}.
\end{definition} 
Formulas $\gnode$ and $h$ in a neighbor criterion specify the (non-fluent and fluent, respectively) criteria on a given neighbor, while formula $\gedge$ specifies the non-fluent criteria on the directed edge from that neighbor to the node in question.

The next component is the ``target criteria'', which are the conditions that a node must satisfy in order to be influenced by its neighbors.  Ideas such as ``susceptibility''~\cite{aral12} can be integrated into our framework via this component.  We represent these criteria with a formula of non-fluent network atoms.  The final component, the ``target'', is simply the label of the target node that is influenced by its neighbors.  Hence, we now have all the pieces that comprise a rule.

\begin{definition}[Rule]
\label{ruleDef}
Given fluent label $L$, natural number $\Delta t$, target criteria $f$ and neighbor criteria\\ $(\gedge,\gnode,h)_\ifl$, a \textit{$\mancalog$ rule} is of the form:
$r = L \tarrow f, (\gedge,\gnode,h)_\ifl$.
We will use the notation $head(r)$ to denote $L$.
\end{definition}
Note that the target (also referred to as the head) of the rule is a single label; essentially, the body of the rule characterizes a set of nodes, and this label is the one that is modified for each node in this set.
More specifically, the rule states that when certain conditions for a node and its neighbors are met, the $\bnd$ bound for the network atom formed with label $L$ on that node changes. Later, in the semantics, we introduce network interpretations, which map components (nodes and edges) of the network to worlds at a given point in time. The rule dictates how this mapping changes in $\Delta t$ time steps.

\begin{definition}[$\mancalog$ Program]
A \textit{program} $P$ is a set of rules, facts, and integrity constraints s.t.\ each non-fluent fact $F \in \calf_{nf}$ appears no more than once in the program.  Let $\mathbf{P}$ be the set of all programs.
\end{definition}

\begin{example}
Following from the running example, supposse $\softtip$ and $\negtip$ are influence functions.  Consider the following rules:
\begin{small}
\begin{eqnarray*}
R_1&=&\watchesA \twoarrow \<\female,[1,1]\>,(\<\strtie,[0.9,1]\>,\true ,\<\watchesA,[0.9,1.0] \>)_\softtip \label{rule1}\\
R_2&=&\watchesB  \onearrow \<\male,[1,1]\>,(\true,\true ,\<\watchesB,[0.8,1.0] \>)_\softtip \label{rule2}\\
R_3&=&\watchesA  \thrarrow \<\male,[1,1]\>,(\true,\<\female,[1,1] \>,\neg\<\watchesA,[0.7,1.0] \>)_\negtip \label{rule3}
\end{eqnarray*}
\end{small}
Rule $R_1$ says that a female node in the network visits page A with a weight specified by the $\mathit{sftTp}$ influence function if at a certain number of her strong ties (with weight of at least $0.9$) visited the page two days ago.
The rest of the rules can be read analogously.
\hfill\smallblacksq
\end{example}


\medskip
\noindent
{\bf Semantics.}
We now introduce our first semantic structure: the \textit{network interpretation}.

\begin{definition}[Network Interpretation]
A \textit{network interpretation} is a mapping of network components to sets of network atoms, $NI : \calg \rightarrow 2^{NA}$.  We will use $\mathbf{NI}$ to denote the set of all network interpretations.
\end{definition}
Note that not all labels will necessarily apply to all nodes and edges in the network.  For instance, certain labels may describe a relationship while others may only describe a property of an individual.  If a given label $L$ does not describe a certain component $c$ of the network, then in a valid network interpretation $NI$, $\< L, [0,1] \> \in NI(c)$.
We define a $\mancalog$ interpretation (simply referred to as ``interpretation'') as follows.

\begin{definition}[Interpretation]
A \textit{$\mancalog$ interpretation} $I$ is a mapping of natural numbers in the interval $[0,\tm]$ to network interpretations,
i.e., $I : \setN \rightarrow \mathbf{NI}$.  Let $\cali$ be the set of all possible interpretations.
\end{definition}



We now need to define {\em satisfaction} of the basic elements by interpretations.
First, we define what it means for an interpretation to satisfy a fact and a rule.

\begin{definition}[Fact Satisfaction]
An interpretation $I$ \textit{satisfies} fact $(a,c):[t_1,t_2]$, written $I \models (a,c):[t_1,t_2]$, iff $\forall t \in [t_1,t_2]$, $I(t)(c)\models a$.
\end{definition}
For non-fluent facts, we introduce the notion of strict satisfaction, which enforces the bound in the interpretation to be set to exactly what the fact dictates.

\begin{definition}[Strict Fact Satisfaction]
In\-ter\-pre\-ta\-tion $I$ \textit{strictly satisfies} fact $(a,c):[t_1,t_2]$ iff $\forall t \in [t_1,t_2]$, $a \in I(t)(c)$.
\end{definition}
Next, we define what it means for an interpretation to satisfy an integrity constraint.

\begin{definition}[IC Satisfaction]
An interpretation $I$ \textit{satisfies} integrity constraint $a \hookleftarrow b$ iff for all $t \in \tau$ and $c \in \calg$, $I(t)(c)\models \neg b \vee a$.
\end{definition}
Before we define rule satisfaction, we require
two auxiliary definitions that are used to define the bound enforced on a label by a given rule, and
the set of time points that are affected by a rule.

\begin{definition}[$\BOUND$ function]
\label{boundDef}
For a given rule $r = L \tarrow f, (\gedge,\gnode,h)_\ifl$, node $v$, and network interpretation $NI$,
$\BOUND(r,v,NI) =$
$\ifl(|\QUAL(v,\gedge,\gnode,h,NI\big)|, |\ELIG(v,\gedge,\gnode,NI)|)$,
where we have
$\ELIG(v,\gedge,\gnode,NI) =$
$\{v' \in V \; | \; NI(v') \models \gnode  \wedge (v',v) \in E\wedge NI((v',v))\models \gedge\}$
and
$\QUAL(v,\gedge,\gnode,h,NI) =$
$\{v' \in \ELIG(v,\gedge,\gnode,NI) \;|\;  NI(v') \models h\}$.
\end{definition}
Intuitively, the bound returned by the function depends on the influence function and the number of qualifying and eligible nodes that influence it.

\begin{definition}[Target Time Set]
For interpretation $I$, node $v$, and rule $r = L \tarrow f, (\gedge,\gnode,h)_\ifl$, the \textit{target time set} of $I,v,r$ is defined as:
$
\mathit{TTS}(I,v,r) = \{ t \in [0,\tm] \; | \;  I(t-\Delta t)(v) \models f\}
$
We also extend this definition to a program $P$, for a given $c \in \calg$ and $L \in \call$, as follows;
$\mathit{TTS}(I,c,L,P) =$
$
\bigcup_{r \in P, \mathit{head}(r)=L}\mathit{TTS}(I,c,r) \cup \{t\in [t_1,t_2] \; | \; (\<L,\bnd\>,c):[t_1,t_2] \in P\}
$
$
\cup \;
\{ t \; | \; \<L, \bnd\> \hookleftarrow b \in P \wedge I(t)(c)\models b\}
$
\end{definition}

\noindent
We can now define satisfaction of a rule by an interpretation.

\begin{definition}
An interpretation $I$ \textit{satisfies} a rule
$
r = L \tarrow f, (\gedge,\gnode,h)_\ifl
$
iff for all $v \in V$ and $t \in \mathit{TTS}(I,v,r)$ it holds that
$
I(t)(v) \models \langle L, \BOUND(r,v,I(t-\Delta t))\rangle
$.
\end{definition}
We now define satisfaction of programs, and introduce {\em canonical interpretations}, in which time points that are not ``targets'' retain information from the last time step.

\begin{definition}[Models and Canonical models]
For interpretation $I$ and program $P$:

\smallskip
\noindent
$I$ is a \textit{model} for $P$ iff it satisfies all rules, integrity constraints, and fluent facts in that program, strictly satisfies all non-fluent facts in the program, and for all $L \in \call,$ $c \in \calg$ and $t \notin \mathit{TTS}(I,c,L,P)$, $\<L,[0,1]\> \in I(c)(t)$.

\smallskip
\noindent
$I$ is a \textit{canonical model} for $P$ iff it satisfies all rules, integrity constraints, and fluent facts in $P$, strictly satisfies all non-fluent facts in $P$, and for all $L \in \call,$ $c \in \calg,$ and $t \notin \mathit{TTS}(I,c,L,P)$, $\<L,[0,1]\> \in I(c)(t)$ when $t=0$ and $\<L,bnd\> \in I(t)(c)$ where $\<L,bnd\> \in I(t-1)(c)$.
\end{definition}


\section{Consistency, Entailment, and Fixpoint Model Computation}
\label{sec:cons-ent-fix}

In this section we discuss consistency and entailment in $\mancalog$ programs, and explore
the use of minimal models towards computing answers to these problems.

\begin{definition}[Consistency and entailment]
A $\mancalog$ program $P$ is \textit{(canonically) consistent} iff there exists a (canonical) model $I$
of $P$.
$P$ (canonically) entails $\mancalog$ fact $F$ iff for all (canonical) models $I$ of $P$, it holds that
$I \models F$.
\end{definition}

Now we define an ordering over models and define the concept of minimal model.  We then show that if we can find a minimal model then we can answer consistency, entailment, and tight entailment queries. We first define a pre-order over interpretations. 
\begin{definition}[Preorder over interpretations, equivalence, and partial ordering]
Given interpretations $I,I'$ we say $I \sqsubseteq^{pre} I'$ iff for all $t, v, L$ if there exists
$\< L, \bnd\> \in I(t)(v)$ then there must exist $\<L, \bnd'\> \in I'(t)(v)$ s.t.\ $\bnd'\subseteq \bnd$.

\smallskip
\noindent
$I, I'$ are \textbf{equivalent} (written $I \sim I'$) iff for all $P \in \mathbf{P}$, $I \models P$ iff $I' \models P$.

\smallskip
\noindent
Given classes of interpretations $[I],[I']$ that are equivalent w.r.t.~$\sim$, we say that $[I]$ precedes $[I']$, written $[I] \sqsubseteq [I']$,
iff $I \sqsubseteq^{pre} I'$.
\end{definition}

\begin{definition}[Minimal Model]
\label{mm-def}
Given program $P$, the \textit{minimal model} of $P$ is a (canonical) interpretation $I$ s.t.\ $I \models P$ and for all (canonical) interpretation $I'$ s.t.\ $I' \models P$,
we have that $I \sqsubseteq I'$.
\end{definition}
We can think of a minimal model of a $\mancalog$ program as the outcome of a multi-attribute process in a complex network that allows us to answer any entailment query.



\medskip
\noindent
{\bf Fixpoint Model Computation.}
We now introduce a fixed-point operator that produces the non-canonical minimal model of a $\mancalog$ program in polynomial time; first, we introduce three preliminary definitions.

\begin{definition}
\label{ibound}
Given program $P$, interpretation $I$, $c \in \calg$, $L \in \call$, and $t\in \tau$, we define functions:

\noindent
$-$ $\FB(P,c,t,L) =$
$\bigcap_{(\<L,\bnd\>,c):[t_1,t_2]\in P\textit{ s.t. }t \in [t_1,t_2]}\bnd$

\noindent
$-$ $\IB(P,c,t,L) =$
$\bigcap_{\<L, \bnd\> \hookleftarrow a \in P \textit{ s.t. }I(t)(c)\models a}\bnd$

\noindent
$-$ $\RB(P,I,v,t,L) =$
$\bigcap_{r\in P\textit{ s.t. }t \in \mathit{TTS}(I,v,L,P)\cap\mathit{TTS}(I,v,r)} \BOUND(r,v,I(t-\Delta t))$.
\end{definition}
We can now introduce the operator.

\begin{definition}[$\Gamma$ Operator]
For a given $\mancalog$ program $P$, we define the operator $\Gamma_P : \cali \rightarrow \cali$ as follows:
For a given $I$, for each $t \in \tau$, $c\in \calg$, and $L \in \call$, add $\<\call, \bnd\>$ to $\Gamma_P(I)(t)(c)$ where $\bnd$ is defined as:
$\bnd = \bnd_{prv}\cap \FB(P,c,t,L) \cap  \IB(P,I,c,t,L) \cap \RB(P,I,c,t,L)$,
where $\<L,\bnd_{prv}\> \in I(t)(c)$.
\end{definition}
It is easy to show that $\Gamma$ can be computed in polynomial time.  Next, we introduce notation for repeated applications of $\Gamma$.
\begin{definition}[Iterated Applications of $\Gamma$]
Given natural number $i > 0$, interpretation $I$, and program $P$, we define $\Gamma_P^i(I)$, the multiple applications
of $\Gamma$:
$\Gamma^i_P(I) = \Gamma_P(I)$ if $i=1$ and $\Gamma^i_P(I) = \Gamma_P(\Gamma^{i-1}_P(I))$ otherwise.
\end{definition}
The iterated $\Gamma$ operator converges after a polynomial number of applications:

\begin{theorem}
\label{gammaPolyConverge}
Given interpretation $I$ and program $P$, there exists a natural number $k$ s.t.\
$\Gamma_P^k(I) = \Gamma_P^{k+1}(I)$, and
$
k \in O(|P|\cdot\dinm \cdot \tm \cdot |E|)
$
where $\dinm$ is the maximum in-degree in the network.
\end{theorem}
In the following, we will use the notation $\Gamma^*_P$ to denote the iterated application of $\Gamma$ after a number of steps sufficient for convergence; Theorem~\ref{gammaPolyConverge} means that we can efficiently compute $\Gamma^*_P$.  We also note that as a single application of $\Gamma$ can be computed in polynomial time, this implies that we can find a minimal model of a $\mancalog$ program in polynomial time.  We now prove the correctness of the operator.  We do this first by proving a key lemma that, when combined with a claim showing that for consistent program $P$, $\Gamma^*_P$ is a model of $P$, tells us that $\Gamma^*_P$ is a minimal model for $P$.  Following directly from this, we have that $P$ is inconsistent iff $\Gamma^*_P=\top$.

\begin{lemma}
\label{boundLemma}
If $I \models P$ and $I' \sqsubseteq I$ then $\Gamma(I') \sqsubseteq I$.
\end{lemma}
\begin{theorem}
\label{minModelGamma}
If program $P$ is consistent then $\Gamma^*_P$ is a minimal model for $P$.
\end{theorem}

\noindent
These results, when taken together, prove that tight entailment and consistency problems for $\mancalog$ can be solved in polynomial time, which is precisely what we set out to
accomplish as part of our desiderata described in Section~\ref{sec:intro}.


\section{An Application in Social Networks: Discussion and Experimental Results}
\label{sec:applications}

An important problem with regard to social networks is to determine group membership of the nodes (individuals).  In particular, we are interested in the problem where some of the individuals in the network have been identified as members of a particular group while the affiliation of the remainder is unknown.  In our work with a major U.S.\ metropolitan police force, we have found this to be an important problem in combating gang violence.
Since in most cases it is considered a criminal offense to simply be in a gang, many gang members deny any type of affiliation upon arrest.  Hence, in order to better understand the dynamics of these criminal organizations, it becomes necessary to use the data at hand to try to identify those with unknown affiliation. One way in which this can be done is by using $\mancalog$ rules that assign a \textit{degree of membership} for each group to each individual with an unknown affiliation; this degree is a number in the interval $[0,1]$ that specifies the confidence that they are in that group.

To address this problem, we propose the following.  Consider a social network of individuals (for the police, this network is created based on co-arrestee data). Each group is assigned a fluent label and, for this problem, only one time point is used.  For each node $i$ that is in a group $g$, we include the fact $(\<g,[1,1]\>,i):[0,0]$ and, for each $g' \neq g$, the fact $(\<g',[0,0]\>,i):[0,0]$.  For all other nodes $j$ we include the fact $(\<g,[0,1]\>,j):[0,0]$ for each group $g$.  We used a simple algorithm (not included due to space constraints) which creates influence functions and rules that assign degrees of membership based on the number of adjacent nodes within a given group.  Then, by using the $\Gamma$ function, we can compute the degree of membership for nodes with an unknown affiliation.

\begin{figure}[t!]
\centering
\includegraphics[width=0.7\textwidth]{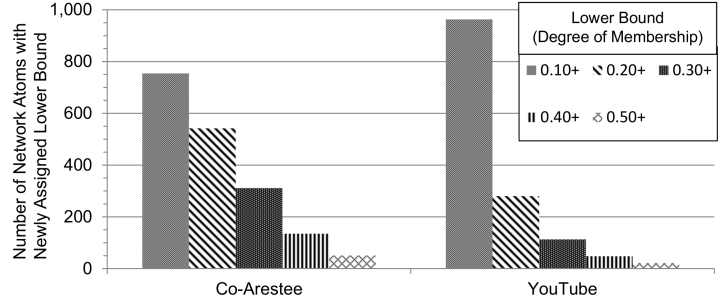}
\caption{Histogram illustrating number of network atoms assigned a lower bound of greater than zero after the convergence of $\Gamma$ (omitting network atoms assigned an initial lower bound of $1.0$).}
\label{degFig}
\end{figure}

\begin{figure}[t!]
\centering
\includegraphics[width=0.5\textwidth]{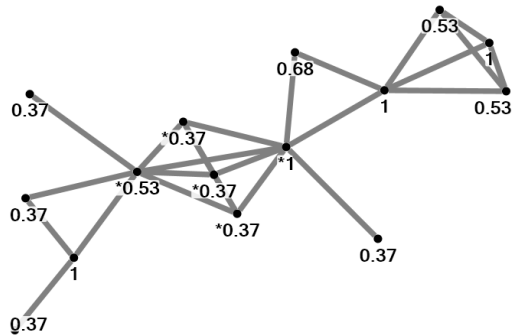}
\caption{Visualization of a subgroup of a faction.  Lower bound of degree of membership is shown.  Core members are denoted with a triangle.}
\label{factionFig}
\end{figure}

\medskip
\noindent
{\bf Implementation and Experimental Results.}
We implemented the $\Gamma$ operator and the computation of its fixed point in Python 2.7.3 in 700 lines of code that leveraged the NetworkX library\footnote{http://networkx.lanl.gov/}.  Additionally, we implemented a rule-learning algorithm and supporting routines in an additional 300 lines of code.  The experiments were run on a computer equipped with an Intel X5677 Xeon Processor operating at 3.46 GHz with a 12 MB Cache running Red Hat Enterprise Linux version 6.1 and equipped with 70 GB of physical memory.

We used two datasets: the previously described gang co-arrestee dataset provided by the police force of a major U.S.\ city, and a network derived from YouTube (based on channel subscriptions)~\cite{yang12}.   The co-arrestee dataset consists of $2,333$ nodes and $3,676$ edges.  The program used for this dataset consists of $58$ rules.  The YouTube dataset consists of $1,134,890$ nodes and $2,987,624$ edges, and we used a program with $47$ rules.  We note that the running time for the convergence of the $\Gamma$ operator for the co-arrestee dataset was $38.41$ seconds, while the running time for the much larger YouTube dataset was $63.6$ hours; though this may be considered a long time, note that it is a one-time computation that allows us to answer many queries once the structure is obtained.

In Figure~\ref{degFig} we illustrate the number of nodes in the network whose lower bound on degree of membership for any of the groups increased after computing the convergence of $\Gamma$.  Note that in our target application, we were able to assign a non-zero degree of membership to several hundred nodes.  With rare exceptions, for the co-arrestee network, nodes were assigned a degree of membership to only one group (gang faction).

In order to get an understanding of the utility of assigning degree of membership, we consider the results of the convergence of the $\Gamma$ operator used as input for some common social network analysis techniques that are likely to aid in police operations.  We examine the sub-graph induced by individuals who had a degree of membership greater than or equal to $0.3$ (a value chosen subjectively given the setup) for a certain gang faction.  We then used the Louvain algorithm~\cite{blondel08} (modularity-maximizing) to identify sub-groups of that faction.  The identification of sub-groups of such factions is useful to police to better understand the structure and dynamics of these organizations in order to improve law enforcement operations.  The sub-graph induced by one such sub-group is shown in Figure~\ref{factionFig}.  Note that the majority of the members in this sub-group have a degree of membership in the faction less than $1$, which means that they were assigned by the $\Gamma$ operator.  This tells us that the sub-group might have been overlooked if degrees of membership were not being computed.  Also, many of the individuals designated as ``core members'' (shown with a triangle in the figure) based on shell decomposition~\cite{Seidman83} were also individuals whose degree of membership was determined by $\Gamma$.  Based on the work of \cite{InfluentialSpreaders_2010}, core members are thought to be key spreaders of information and thus also of interest for policing operations, particularly with regard to gathering intelligence on the sub-group in question.



\section{Conclusions and Future Work}
\label{sec:conc}

In this paper, we presented the $\mancalog$ language for describing multi-attribute networks and cascades.  We started by recalling seven criteria in the form of desiderata for such a formalism, and showed that $\mancalog$ meets all of them; to the best of our knowledge, this has not been accomplished by any previous model in the literature.  We also implemented this language and applied it to the degree of membership problem in social networks and showed how the results can aid in real-world law enforcement operations.  We also note that $\mancalog$ is the first language of its kind to consider network structure in the semantics, potentially opening the door for algorithms that leverage features of network topology to more efficient query answering algorithms.

Currently, we are looking at other applications of $\mancalog$ as well as methods to learn rules that describe diffusion processes in social networks.  In the near future, we shall also explore various types of queries that have been studied in the literature, such as finding nodes of maximum influence, identifying nodes that cause a cascade to spread more quickly, and identifying nodes that can be influenced in order to halt a cascade.

\smallskip
\noindent
{\bf Acknowledgments.} This work was supported by
ARO project 2GDATXR042 and
UK EPSRC grant EP/J008346/1 (``PrOQAW'').


\bibliographystyle{acmtrans}
\bibliography{network}

\begin{thebibliography}{}

\bibitem[\protect\citeauthoryear{Aral and Walker}{Aral and
  Walker}{2012}]{aral12}
{\sc Aral, S.} {\sc and} {\sc Walker, D.} 2012.
\newblock {Identifying Influential and Susceptible Members of Social Networks}.
\newblock {\em Science\/}~{\em 337,\/}~6092, 337--341.

\bibitem[\protect\citeauthoryear{Blondel, Guillaume, Lambiotte, and
  Lefebvre}{Blondel et~al\mbox{.}}{2008}]{blondel08}
{\sc Blondel, V.}, {\sc Guillaume, J.}, {\sc Lambiotte, R.}, {\sc and} {\sc
  Lefebvre, E.} 2008.
\newblock {Fast unfolding of communities in large networks}.
\newblock {\em Journal of Statistical Mechanics: Theory and Experiment\/}~{\em
  2008}, P10008.

\bibitem[\protect\citeauthoryear{Broecheler, Shakarian, and
  Subrahmanian}{Broecheler et~al\mbox{.}}{2010}]{bss10}
{\sc Broecheler, M.}, {\sc Shakarian, P.}, {\sc and} {\sc Subrahmanian, V.~S.}
  2010.
\newblock A scalable framework for modeling competitive diffusion in social
  networks.
\newblock In {\em Proc.\ of SocialCom}. IEEE.

\bibitem[\protect\citeauthoryear{Granovetter}{Granovetter}{1978}]{Gran78}
{\sc Granovetter, M.} 1978.
\newblock Threshold models of collective behavior.
\newblock {\em The American Journal of Sociology\/}~{\em 83,\/}~6, 1420--1443.

\bibitem[\protect\citeauthoryear{Kempe, Kleinberg, and Tardos}{Kempe
  et~al\mbox{.}}{2003}]{kleinberg}
{\sc Kempe, D.}, {\sc Kleinberg, J.}, {\sc and} {\sc Tardos, E.} 2003.
\newblock Maximizing the spread of influence through a social network.
\newblock In {\em Proc.\ of KDD '03}. ACM, 137--146.

\bibitem[\protect\citeauthoryear{Kitsak, Gallos, Havlin, Liljeros, Muchnik,
  Stanley, and Makse}{Kitsak et~al\mbox{.}}{2010}]{InfluentialSpreaders_2010}
{\sc Kitsak, M.}, {\sc Gallos, L.~K.}, {\sc Havlin, S.}, {\sc Liljeros, F.},
  {\sc Muchnik, L.}, {\sc Stanley, H.~E.}, {\sc and} {\sc Makse, H.~A.} 2010.
\newblock {Identification of influential spreaders in complex networks}.
\newblock {\em Nat Phys\/}~{\em 6,\/}~11 (Nov.), 888--893.

\bibitem[\protect\citeauthoryear{Lieberman, Hauert, and Nowak}{Lieberman
  et~al\mbox{.}}{2005}]{liebermanEvolutionary2005}
{\sc Lieberman, E.}, {\sc Hauert, C.}, {\sc and} {\sc Nowak, M.~A.} 2005.
\newblock Evolutionary dynamics on graphs.
\newblock {\em Nature\/}~{\em 433,\/}~7023, 312--316.

\bibitem[\protect\citeauthoryear{Schelling}{Schelling}{1978}]{schelling}
{\sc Schelling, T.~C.} 1978.
\newblock {\em Micromotives and Macrobehavior}.
\newblock W.W. Norton and Co.

\bibitem[\protect\citeauthoryear{Seidman}{Seidman}{1983}]{Seidman83}
{\sc Seidman, S.~B.} 1983.
\newblock Network structure and minimum degree.
\newblock {\em Social Networks\/}~{\em 5,\/}~3, 269 -- 287.

\bibitem[\protect\citeauthoryear{Shakarian, Simari, and Schroeder}{Shakarian
  et~al\mbox{.}}{2013}]{mancalogAAMAS13}
{\sc Shakarian, P.}, {\sc Simari, G.~I.}, {\sc and} {\sc Schroeder, R.} 2013.
\newblock {MANCaLog}: A logic for multi-attribute network cascades.
\newblock In {\em Proc.\ of AAMAS-2013}.

\bibitem[\protect\citeauthoryear{Shakarian, Subrahmanian, and Sapino}{Shakarian
  et~al\mbox{.}}{2010}]{snops}
{\sc Shakarian, P.}, {\sc Subrahmanian, V.~S.}, {\sc and} {\sc Sapino, M.~L.}
  2010.
\newblock {Using Generalized Annotated Programs to Solve Social Network
  Optimization Problems}.
\newblock In {\em Proc.\ of ICLP (tech.\ comm.)}.

\bibitem[\protect\citeauthoryear{Sood, Antal, and Redner}{Sood
  et~al\mbox{.}}{2008}]{sood08}
{\sc Sood, V.}, {\sc Antal, T.}, {\sc and} {\sc Redner, S.} 2008.
\newblock Voter models on heterogeneous networks.
\newblock {\em Physical Review E\/}~{\em 77,\/}~4, 041121.

\bibitem[\protect\citeauthoryear{Yang and Leskovec}{Yang and
  Leskovec}{2012}]{yang12}
{\sc Yang, J.} {\sc and} {\sc Leskovec, J.} 2012.
\newblock Defining and evaluating network communities based on ground-truth.
\newblock In {\em Proceedings of the ACM SIGKDD Workshop on Mining Data
  Semantics}. MDS '12. ACM, New York, NY, USA, 3:1--3:8.

\end{thebibliography}
\label{lastpage}





\end{document}